\def\year{2022}\relax
\documentclass[letterpaper]{article} 
\usepackage{aaai22}  
\usepackage{times}  
\usepackage{helvet}  
\usepackage{courier}  
\usepackage[hyphens]{url}  
\usepackage{graphicx} 
\usepackage{natbib}  
\usepackage{caption} 
\DeclareCaptionStyle{ruled}{labelfont=normalfont,labelsep=colon,strut=off} 
\usepackage{algorithm}
\usepackage{algorithmic}

\usepackage{amsmath}
\usepackage{amssymb}
\usepackage{newfloat}
\usepackage{listings}
\floatstyle{ruled}
\newfloat{listing}{tb}{lst}{}
\floatname{listing}{Listing}
\title{Learning Quality-aware Representation for Multi-person Pose Regression}

\author {
    Yabo Xiao,\textsuperscript{\rm 1,}\footnote{Equal contribution. $^{\dagger}$ Corresponding author.}
    Dongdong Yu, \textsuperscript{\rm 2,}\footnotemark[1]
    Xiaojuan Wang, \textsuperscript{\rm 1,}\footnotemark[2]
    Lei Jin, \textsuperscript{\rm 1}
    Guoli Wang, \textsuperscript{\rm 3}
    Qian Zhang \textsuperscript{\rm 4}
    }

\affiliations {
    \textsuperscript{\rm 1} Beijing University of Posts and Telecommunications\\
    \textsuperscript{\rm 2} ByteDance Inc.\\
    \textsuperscript{\rm 3} Tsinghua University\\
    \textsuperscript{\rm 4} Horizon Robotics\\
    \{xiaoyabo, wj2718, jinlei\}@bupt.edu.cn, yudongdong@bytedance.com, wangguoli1990@mail.tsinghua.edu.cn, qian01.zhang@horizon.ai
}

\begin{document}

\maketitle

\begin{abstract}

Off-the-shelf single-stage multi-person pose regression methods generally leverage the instance score (i.e., confidence of the instance localization) to indicate the pose quality for selecting the pose candidates. We consider that there are two gaps involved in existing paradigm:~1) The instance score is not well interrelated with the pose regression quality.~2) The instance feature representation, which is used for predicting the instance score, does not explicitly encode the structural pose information to predict the reasonable score that represents pose regression quality. To address the aforementioned issues, we propose to learn the pose regression quality-aware representation. Concretely, for the first gap, instead of using the previous instance confidence label (e.g., discrete \{1,0\} or Gaussian representation) to denote the position and confidence for person instance, we firstly introduce the Consistent Instance Representation (CIR) that unifies the pose regression quality score of instance and the confidence of background into a pixel-wise score map to calibrates the inconsistency between instance score and pose regression quality. To fill the second gap, we further present the Query Encoding Module (QEM) including the Keypoint Query Encoding (KQE) to encode the positional and semantic information for each keypoint and the Pose Query Encoding (PQE) which explicitly encodes the predicted structural pose information to better fit the Consistent Instance Representation (CIR). By using the proposed components, we significantly alleviate the above gaps. Our method outperforms previous single-stage regression-based even bottom-up methods and achieves the state-of-the-art result of 71.7 AP on MS COCO test-dev set.

\end{abstract}

\section{Introduction}
Given an input RGB image, multi-person pose estimation aims to detect the keypoint positions for all persons. With the prevalence of deep learning \cite{deng2009imagenet,newell2016stacked,ren2015faster}, it has attracted much attention since it plays as an important role in many computer vision tasks such as pose tracking \cite{xiao2018simple,DongdongYu2018MultipersonPE}, activity recognition \cite{li2019actional,shi2019two}, human re-identification and so on. 

\begin{figure}
\begin{center}

\includegraphics[height=0.613\columnwidth,width=1\columnwidth, trim=110 40 110 40]{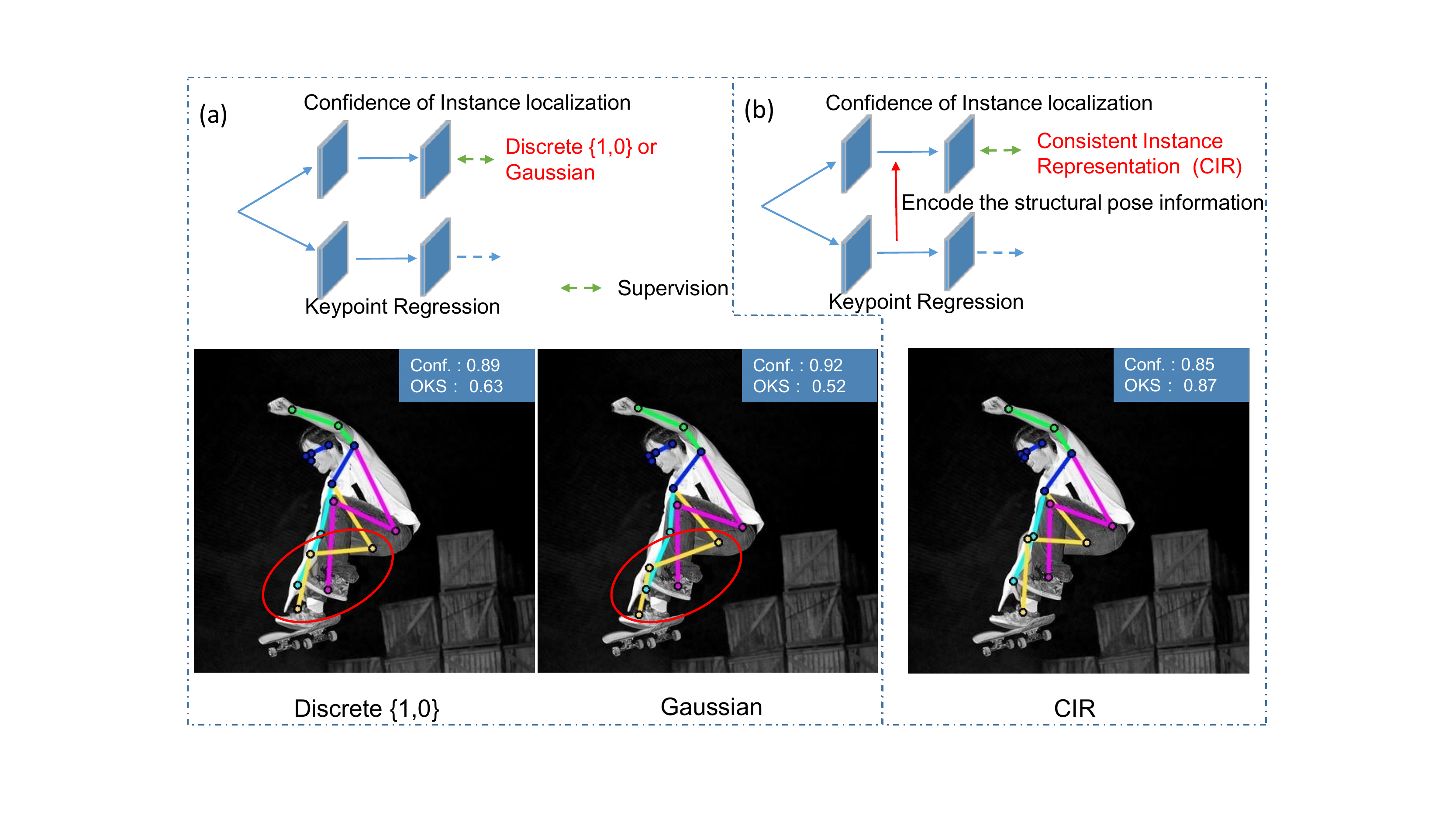}
\end{center}
\caption{(a) Off-the-shelf single-stage pose regression pipeline and the predicted human pose with high instance score (Conf.) and inferior pose quality (wrong
localization for leg area) when using discrete \{1,0\} or Gaussian kernel to supervise the instance score. (b) Our proposed pipeline predicts the pose with high instance score (Conf.) and superior pose regression quality. The pose regression quality is quantified as OKS.
}
\label{fig:image1}
\vspace{-3mm}
\end{figure}

Most existing multi-person pose estimation methods follow top-down pipeline \cite{chen2018cascaded,sun2018integral,yu2018deep,xiao2018simple, su2019multi} and bottom-up pipeline \cite{cao2017realtime, cheng2020higherhrnet,kreiss2019pifpaf,papandreou2018personlab,newell2017associative}. The top-down methods firstly detect the region of person instance via object detector \cite{cai2018cascade,law2018cornernet,tian2019fcos}, then perform single person pose estimation on the cropped human body regions. Generally, the top-down pipeline is limited by the detection-first paradigm which leads to high computation and memory cost. The bottom-up methods firstly locate the keypoints of all persons in an image simultaneously and then assign the keypoints to individuals via a grouping process. However, the additional grouping process is computationally complex.

By contrast, recent practices \cite{nie2019single,tian2019directpose} deliver a single-stage solution via pixel-wise keypoint regression, which is more straightforward and bypasses the above drawbacks of previous methods. In particular, it decomposes the pose estimation task into person instance localization and corresponding keypoint regression. For example, SPM \cite{nie2019single} leverages the Gaussian kernel to indicate the confidence of instance localization and proposes a hierarchical pose representation to regress joints. DirectPose \cite{tian2019directpose} employs the discrete instance confidence label \{1,0\} to denote the position and confidence of the person instance and presents a KPAlign scheme to locate the keypoints. However, we consider that there are two gaps in the above practices: 1) As shown in Figure \ref{fig:image1}(a), the instance score supervised by prior instance confidence label (e.g., discrete \{1,0\} or Gaussian representation) is not well correlated with the pose regression quality. 2) The instance feature representation used for predicting the instance score does not explicitly encode the predicted structural pose information, thus is hard to accurately estimate the pose quality score.


To alleviate the above issues, in contrast to previous methods that leverage the discrete \{1,0\} or Gaussian kernel to denote the confidence of instance localization, we firstly propose to construct the Consistent Instance Representation (CIR) that unifies the pose regression quality score of person instance and the confidence of background into a pixel-wise score map to fill the first gap. Thus, the CIR is able to denote the position of person instance and corresponding pose regression quality simultaneously. Furthermore, to address the second gap, we introduce the Query Encoding Module (QEM) including the Keypoint Query Encoding (KQE) and the Pose Query Encoding (PQE). Specifically, we utilize the Keypoint Query Encoding (KQE) to encode the positional and semantic information of each keypoint for precisely regression, as well as leverage the Pose Query Encoding (PQE) to explicitly encode the predicted structural pose information into instance feature via positional information of each keypoint query. As shown in Figure \ref{fig:image1}(b), our proposed approach is able to alleviate the inconsistency between instance score and pose regression quality via the proposed Consistent Instance Representation (CIR) and Query Encoding Module (QEM).

The main contributions in our paper can be summarized as follows:

\begin{itemize}
\item We propose the Consistent Instance Representation(CIR) that unifies the pose regression quality score of person instance and the presence confidence of background into a single pixel-wise score map, which alleviates the inconsistency between instance score and pose regression quality.

\item We further introduce a Query Encoding Module (QEM) including Keypoint Query Encoding (KQE) to encode the positional and semantic information of each keypoint for precisely regression and Pose Query Encoding(PQE) which explicitly involves the predicted structural pose information into instance feature via the positions of keypoint queries predicted by KQE.



\item Based on the proposed CIR and QEM, our approach outperforms previous single-stage even bottom-up methods and achieves the state-of-the-art performance with 71.7 AP on MS COCO test-dev set. To our best knowledge, our paper is the first to attempt to fill the aforementioned gaps in the single-stage pose regression paradigm.


\end{itemize}

\begin{figure*}
\begin{center}
\includegraphics[height=0.969\columnwidth,width=1.9\columnwidth, trim=25 35 25 35]{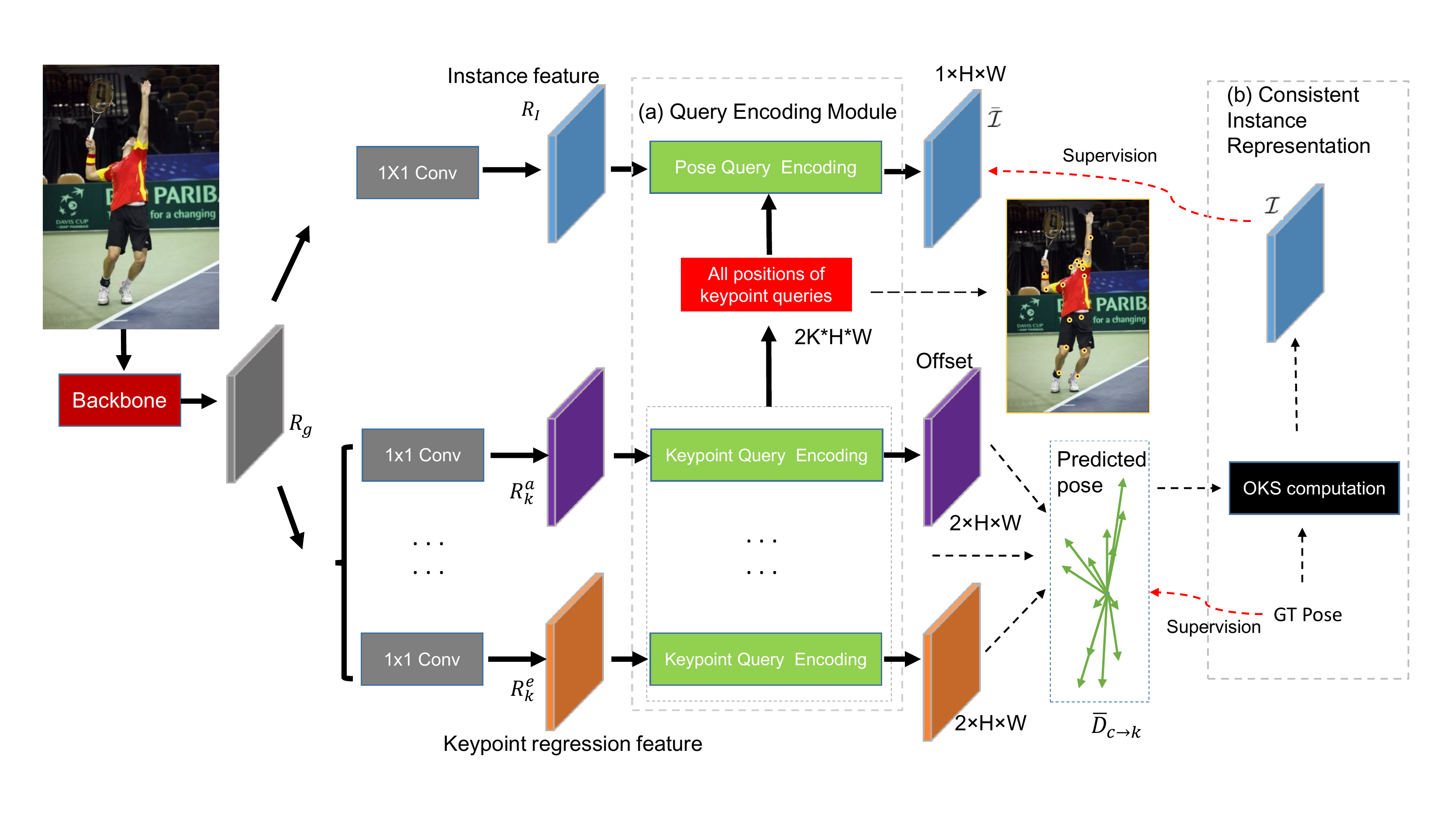}
\end{center}
\caption{ The schematic diagram of our proposed network which uses the multi-branch structure to separately regress the keypoint. Note that $R_k$ represents the different keypoint features (e.g., $R_k^{a}$ for ankle, $R_k^{e}$ for elbow).~(a) The proposed Query Encoding Module including Keypoint Query Encoding which encodes the position and sematic information of each keypoint for precisely regression and Pose Query Encoding that merges the structured pose information into instance feature. (b) The Consistent Instance Representation that leverages the pose regression quality score to denote the confidence of instance localization.~ The position of keypoint queries are visualized by yellow circle. K is the number of keypoints queries.}
\label{fig:image2}
\end{figure*}


\section{Related Work}
In this section, we will review three aspects related to our paper consist of quality estimation, top-down and bottom-up methods as well as single-stage pose regression.


{\bf Quality estimation.} Recently, quality estimation is applied in many vision tasks. For instance, IOU-Net \cite{jiang2018acquisition} adds a parallel branch to predict the IOU between each detected bounding box and the matched ground-truth, which improves the NMS procedure by preserving accurately localized bounding boxes. Mask Scoring RCNN \cite{huang2019mask} proposes to learn the quality of the predicted instance masks via a network block which takes the instance feature and the corresponding predicted mask together to regress the mask IOU. FCOS \cite{tian2019fcos} presents a centerness branch to suppress the low-quality detections produced by the locations far away from the object center. The above methods add an additional branch to predict the quality score and the supervision is only assigned for positive samples. Moreover, the classfication score and quality score are multiplied for conducting NMS process. By contrast, we leverage a single branch to learn the proposed Consistent Instance Representation (CIR), which avoids the burden of additional quality estimation branch and the unreliable bias caused by multiplying the instance score with improper quality score.

{\bf Top-down and bottom-up methods.} Most existing multi-person pose estimation works can be categorized into top-down and bottom-up methods. Top-down methods \cite{chen2018cascaded,fang2017rmpe,sun2018integral,sun2019deep} firstly detect and crop the person region from the image and then predict the single person pose. HRNet~\cite{sun2019deep} maintains high-resolution representations and repeatedly aggregates multi-resolution representations to obtain reliable high-resolution representations. SPCNet \cite{YaboXiao2020SPCNetSpatialPA} proposes to incorporate a Dilated Hourglass Module and a Selective Information Module into Hourglass-style network which preserves high spatial resolution and adaptively assembles the multi-level features for detecting the single-person keypoints. Bottom-up methods \cite{cao2017realtime,cheng2020higherhrnet,newell2017associative} firstly detect the all keypoints in the image and assign the keypoints to individuals via a heuristic grouping process. Associative Embedding \cite{newell2017associative} proposes to predict the keypoint heatmap and tag map simultaneously, and then groups the keypoints with similar tags into individuals. HigherHRNet\cite{cheng2020higherhrnet} presents a Higher-Resolution Network to learn high-resolution feature pyramids to better detect the keypoints of the persons with various scales, and follows the AE \cite{newell2017associative} to group the keypoints. Both top-down and bottom-up methods need a additional stage to associate the person instance with corresponding keypoints.


{\bf Single-stage pose regression.} Recent practices provide a single-stage solution via pixel-wise instance localization and corresponding keypoint regression. They have draw much attention since its compact and end-to-end pipeline. For instance, DirectPose \cite{tian2019directpose} proposes a keypoint alignment (KPAlign) module to overcome the misalignment between the features and the predictions. SPM \cite{nie2019single} proposes a hierarchical Structured Pose Representation according to the body structure to regress accumulative short-range offsets instead of directly regressing long-range offset. PointsetNet \cite{wei2020point} performs regression from a set of points placed at more advantageous positions which provide informative features and task-specific initializations. However, the above methods concentrates on how to accurately regress the keypoint while ignore the two gaps mentioned above. In this paper, we propose the Consistent Instance Representation and the Query Encoding Module to involve the structural pose information into the instance confidence label and instance feature representation simultaneously to fill the two gaps.





\section{Methods}
In this section, we firstly review the formulation of single-stage pixel-wise pose regression. Then, we describe the proposed Consistent Instance Representation (CIR). Finally, we elaborate on the proposed Query Encoding Module (QEM) including Keypoint Query Encoding (KQE) and Pose Query Encoding (PQE).


\subsection{Single-stage Pose Regression formulation}\label{subsection3.1} 

Single-stage multi-person pose estimation methods generally formulate this task as pixel-wise person instance localization and corresponding keypoint regression. It firstly encodes the input image $\mathit{I}$ to produce the general feature representation via the backbone, which is formulated as $\mathit{R_{g}} = \mathcal{F}_{backbone}(\mathit{I}) \in \mathbb{R}^{C \times H \times W} $. Then the followed two parallel sub-branch are employed to perform pixel-wise person instance localization and corresponding keypoint offset regression. Concretely, one is to convert the ${\bf \mathit{R_{g}}}$ to instance representation $\mathit{R_{I}}$, in which each pixel embedding is used to represent an instance and predict the corresponding instance score. The other is to transform $\mathit{R_{g}}$ to keypoint regression representation $\mathit{R_{k}}$, from which the pixel embedding is used to regress the corresponding keypoint displacements. During inference, the regressed pose with high instance score is selected as pose candidates to evaluate the performance.




We consider the above formulation exists two gaps: 1) The instance score is used to select the pose candidates while it is not well correlated with the pose regression quality. 2) The pixel embedding in instance feature $\mathit{R_{I}}$ does not explicitly encode the corresponding predicted pose information to estimate the reasonable pose regression quality score. Thus we propose the Consistent Instance Representation and Query Encoding Module to attempt to fill the above gaps.  




\subsection{Consistent Instance Representation}\label{subsection3.1} 

To address the inconsistency between the instance score and pose regression quality caused by using the prior instance confidence label (discrete \{1,0\} and Gaussian kernel) to denote the confidence of instance localization, we construct the Consistent Instance Representation (CIR) that leverages the pose regression quality score to indicate the position and confidence of person instance.

\begin{figure}
\begin{center}
\includegraphics[height=0.51\columnwidth,width=\columnwidth]{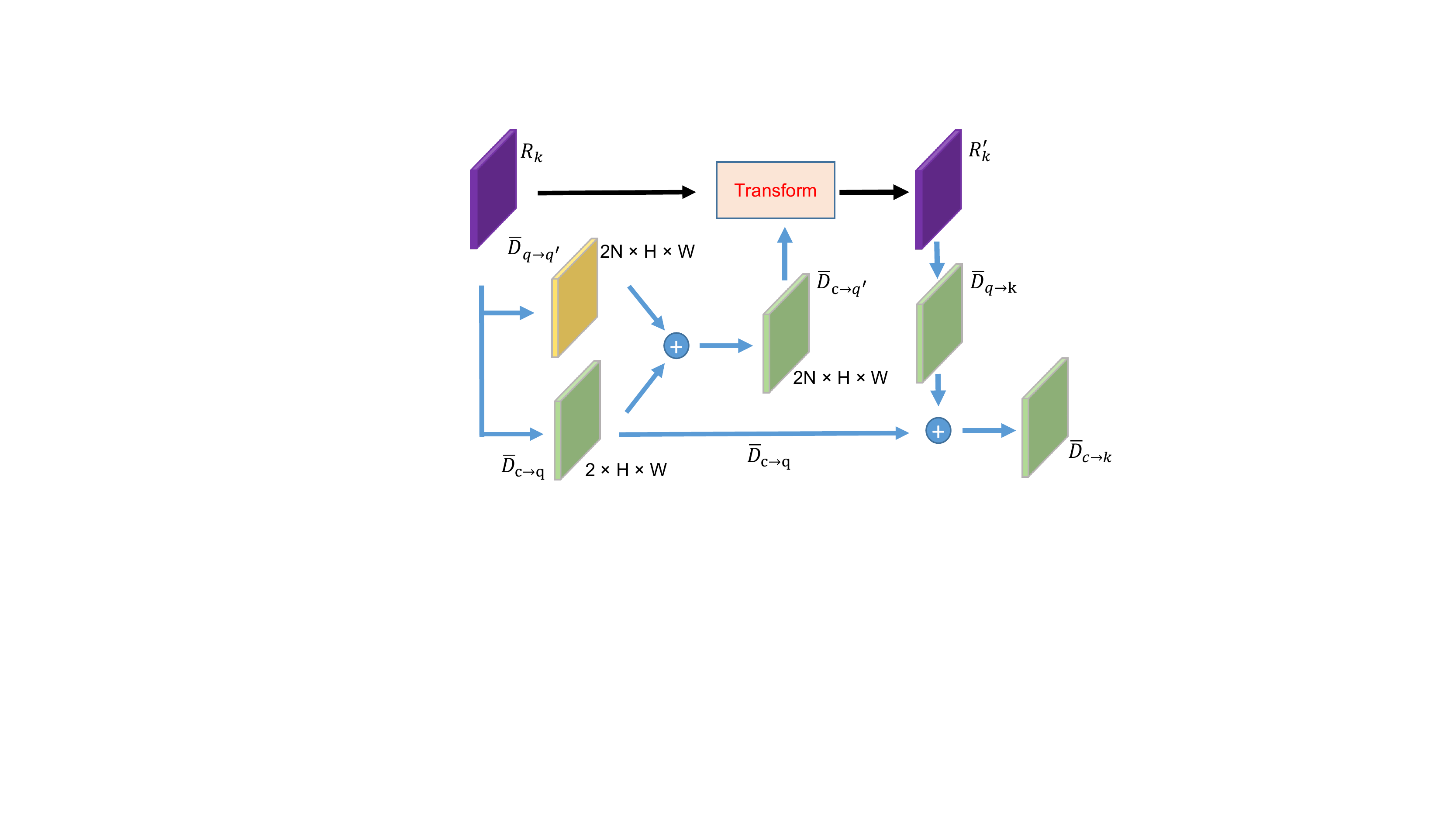}
\end{center}
\caption{The architecture of Keypoint Query Encoding. Transform refers to extract and aggregate the feature vectors of semantic points $q_{'}$. N is the number of semantic points.}
\label{fig:image3}
\end{figure}

We leverage Object Keypoint Similarity~(OKS) \cite{lin2014microsoft} between the predicted human pose and the corresponding ground-truth pose to quantify the pose regression quality. In particular, the Object Keypoint Similarity~(OKS) of pose $\mathcal{P}$ is formulated as follows:

\begin{equation}
   {OKS}_{\mathcal{P}}= \frac{\sum_{i}exp(\frac{-{d}_{\mathcal{P}, i}^{2}}{2{s}_{\mathcal{P}}^{2}{k}_{i}^{2}}) \delta(\upsilon_{\mathcal{P},i}>0)}    {\sum_{i}\delta(\upsilon_{\mathcal{P},i}>0)},
\end{equation}
where $d_{\mathcal{P},i}$ is the Euclidean distance between the i-th predicted keypoint location and the ground-truth one of pose $\mathcal{P}$, $\upsilon_{\mathcal{P},i}$ indicates visible or not for the i-th keypoint of pose $\mathcal{P}$, $s_{\mathcal{P}}$ refers to the instance scale of $\mathcal{P}$, and $k_{i}$ is a constant to control falloff for the i-th keypoints.

The Consistent Instance Representation is a pixel-wise score map denoted as $\mathcal{I}$. $\mathcal{I}(x,y)$ refers to the score at position (x, y), which is formulated as:

\begin{equation} \label{formula:oksmap}
\mathcal{I}(x,y)= 
\left\{
\begin{array}{l}
\begin{aligned}
OKS(\mathcal{\bar P}_{(x,y)}, \mathcal{P}_{n}) & & if~~(x,y) \in \Omega_{n} \\
0    & & else  ,\\

\end{aligned}
\end{array}
\right.
\end{equation}

\noindent where $\mathcal{\bar P}_{(x,y)}$ is the predicted pose at the position (x,y)$\in \Omega_{n}$ and $\mathcal{P}_{n}$ refers to ground-truth pose of the n-th human instance, $\Omega_{n}$ is neighboring area around the n-th human instance center $(x_{n}^{c},y_{n}^{c})$, which is formulated as $\{(x,y)~|~\sqrt{[(x,y)-(x_{n}^{c},y_{n}^{c})]^{2}}<\gamma \}$, $\gamma$ indicates the radius of neighboring area. The Consistent Instance Representation $\mathcal{I}$ $\in$ [0,1] is able to discriminate person instance and background, in which scalar 0 indicates the confidence of background and the others indicate the person instance position and corresponding pose regression quality score.


\begin{figure}
\begin{center}
\includegraphics[height=0.516\columnwidth,width=0.9\columnwidth]{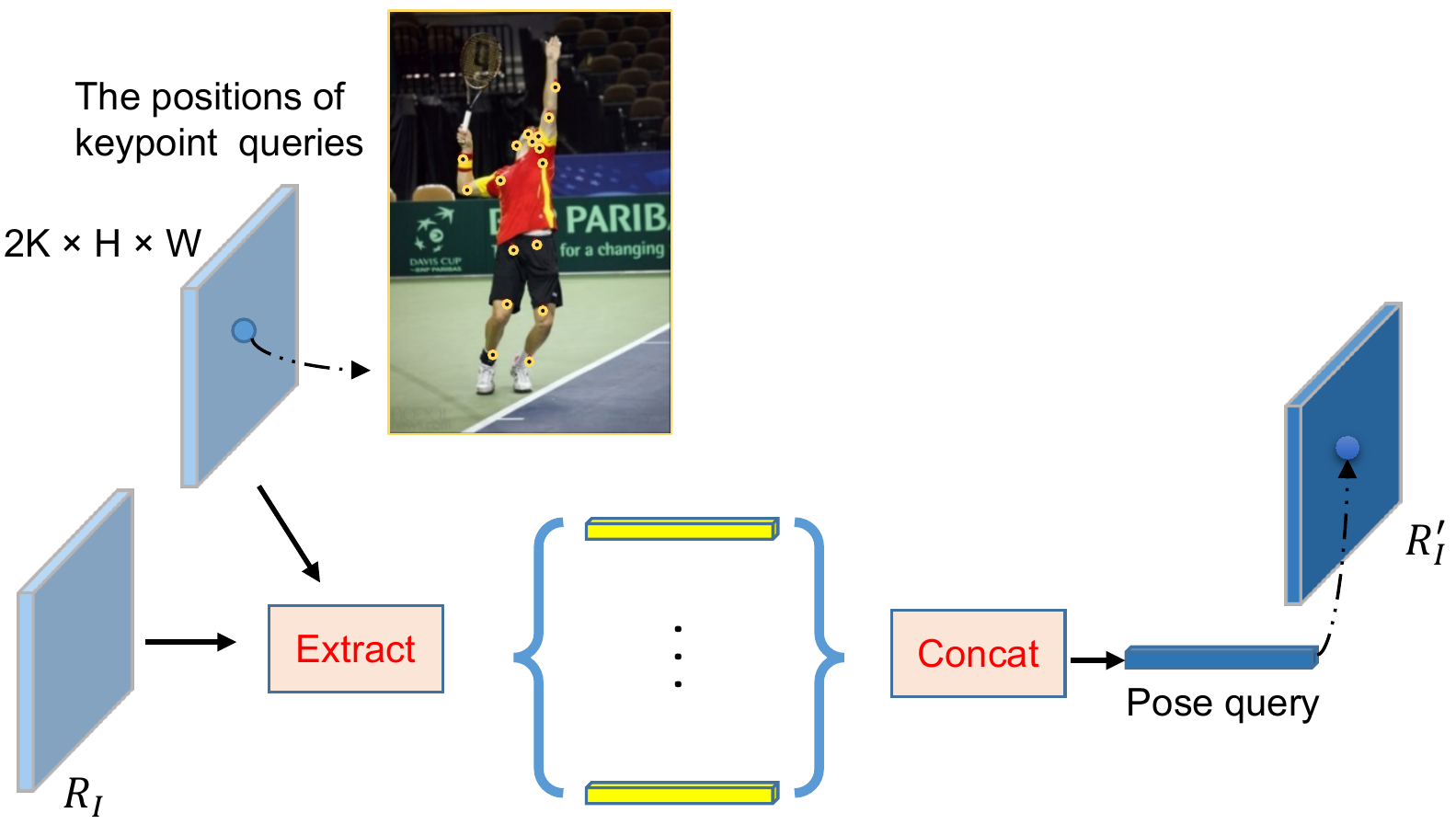}
\end{center}
\caption{The architecture of Pose Query Encoding.}
\label{fig:image4}
\end{figure}


As shown in Figure \ref{fig:image2}(b), we predict a score map $\bar{\mathcal{I}}$ to learn the Consistent Instance Representation $\mathcal{I}$ and employ the pixel-wise weighted L2 loss to penalize the predictions as follows:

\begin{equation}
   {L}_{\mathcal{I}}=  W ~ * \parallel \bar{\mathcal{I}}- \mathcal{I} \parallel ^{2},
\label{formula:loss_oks}
\end{equation}

\noindent where ${W}$ denotes the pixel-wise weight map, the weight of the human central
area is set to 1 and the background region is 0.1. $*$ refers to pixel-wise multiplication operation.


\subsection{Query Encoding Module}\label{subsection3.2} 

In order to encode the predicted structural pose information for predicting reasonable pose score, we propose the Query Encoding Module (QEM) including Keypoint Query Encoding (KQE) and Pose Query Encoding (PQE), as shown in Figure\ref{fig:image2}(a). Keypoint Query Encoding (KQE) is presented to encode the positional and semantic information of each keypoint. Pose Query Encoding (PQE) is introduced to encode the structural pose information into instance feature via the each keypoint query's location predicted by KQE.



\begin{table*}
\begin{center}

\resizebox{2.0\columnwidth}{!}{

\begin{tabular}{l|c|c|c|ccccc|c}
\hline
Methods& Params &Input size & GFLOPs & $AP$ & $AP_{50}$ & $AP_{75}$ & $AP_{M}$ & $AP_{L}$ &  $AR$ \\

\hline

Personlab~\cite{papandreou2018personlab}&68.7 & 1401 &405.5 &66.5 & 86.2 & 71.9 & 62.3 & 73.2 & 70.7 \\
PifPaf~\cite{kreiss2019pifpaf}&- & - &- &67.4&- &- & -& -& - \\

CenterNet-DLA~\cite{zhou2019objects}& -&512 &-&58.9& - &- & -& -& - \\
CenterNet-HG~\cite{zhou2019objects}& -& 512 &- &64.0 & - & - & - & - & - \\
HrHRNet-W32\cite{cheng2020higherhrnet}&28.5 &512&47.9 & 67.1 & 86.2 & 73.0 & - & - & -  \\
HrHRNet-W48\cite{cheng2020higherhrnet}& 63.8 &640&154.3 & 69.9 & 87.2 & 76.1 & - & - & -   \\
DEKR-W32\cite{geng2021bottom}&29.6 &512& 45.4&68.0&86.7&74.5&62.1&77.7&73.0\\
DEKR-W48\cite{geng2021bottom}&65.7 &640&141.5 &71.0&88.3&77.4&66.7&78.5&76.0\\

\hline
Ours~(HRNet-W32)& 29.7&512&46.4 &69.8 & 88.1 & 76.2 & 63.8& 78.9 & 73.8   \\
\bf{Ours~(HRNet-W48)}&65.8 &640&143.4 &{\bf72.4} & {\bf89.1} & {\bf79.0} & {\bf67.3} & {\bf80.4} & {\bf76.4}\\
\hline

\end{tabular}
}
{\caption{Comparisons with the previous state-of-the-art methods on the COCO mini-val set (single-scale testing). }\label{tab:val}}

\end{center}
\end{table*}

{\bf Keypoint Query Encoding.} We use multi-branch structure to perform separated keypoint regression, in which each branch follows the same design. We leverage the general feature $R_{g}$ to produce the separated keypoint representation $R_{k}$ (e.g. nose representation $R_{k}^{n}$, ankle representation $R_{k}^{a}$, elbow representation $R_{k}^{e}$, wrist representation ${R}_{k}^{w}$) via the separated 1$\times$1 convolutional layer and define a keypoint query for each keypoint, which encodes the positional and semantic information to precisely localize the keypoints. 

Specifically, we take a pixel position $c$ of instance central area as example to illustrate the Keypoint Query Encoding. The whole process is decomposed into three steps. As shown in Figure \ref{fig:image3}, first, we predict a displacement $\bar{\bf D}_{c\rightarrow q}$ from center $c$ to dynamically locate the keypoint query $q$ via the separated keypoint representation $R_{k}$. Second, owing to keypoint query is a single position and thus limited to encode the sufficient semantic information of corresponding keypoint, we further extract and aggregate the features of N points (named semantic point $q_{'}$) to enrich the semantic information of keypoint query via bilinear interpolation. The semantic points $q_{'}$ are located by regressing N displacements $\bar{\bf D}_{q\rightarrow q^{'}}$ based on the position of keypoint query $q$. Obtaining the transformed feature ${ R}_{k}^{'}$ is formualted as : $R_{k}^{'}(c) = \sum_{n=1}^N\{{ R}_{k}(c+ \bar{\bf D}_{c\rightarrow q} + \bar{\bf D}_{q\rightarrow q_n^{'}})\}$, where N is the number of semantic points. Thus, the transformed feature $R_{k}^{'}$ is considered to encode the sufficient positional and semantic information for each keypoint. Finally, we regress the displacements $\bar{\bf D}_{q\rightarrow k}$ from the keypoint query to corresponding keypoint via the transformed feature $R_{k}^{'}$ to precisely locate the keypoint. The displacements $\bar{\bf D}_{c\rightarrow k}$ from center to keypoints is formulated as:
\begin{equation}
\bar{\bf D}_{c\rightarrow k}^{i} = \bar {\bf D}_{c\rightarrow q}^{i} + \bar{\bf D}_{q\rightarrow k}^{i},
\end{equation}
where i refers to the i-th keypoint of human instance.




We construct a pixel-wise dense offset map ${\bf D}_{c\rightarrow k}$ as ground-truth to penalize the predicted $\bar{\bf D}_{c\rightarrow k}$. ${\bf D}_{c\rightarrow k}^{i}$ indicates the x-y offsets for the i-th keypoints of the person, takes the following form:

\begin{equation} \label{formula:target_offset}
{\bf D}_{c\rightarrow k}^{i}(x,y)= 
\left\{
\begin{array}{l}
\begin{aligned}

(x,y) - (x_{n}^{i},y_{n}^{i}) & & if~~(x,y) \in \Omega_{n} \\
0 & & else  ,\\

\end{aligned}
\end{array}
\right.
\end{equation}
where $\Omega_{n}$ is adjacent area around the n-th instance center $(x_{n}^{c},y_{n}^{c})$, which is formulated as $\{(x,y)~|~\sqrt{[(x,y)-(x_{n}^{c},y_{n}^{c})]^{2}}<\gamma \}$. $(x_{n}^{i},y_{n}^{i})$ indicates the coordinates of the i-th keypoint for the n-th person. The loss function is smooth L1 as follows:

\begin{equation}
{L}_{{\bf D}_{c\rightarrow k}}=  Smooth L1({\bf D}_{c\rightarrow k}, \bar{\bf D}_{c\rightarrow k}).
\label{formula:loss_displacement}
\end{equation}


{\bf Pose Query Encoding.} Based on the keypoint queries, we introduce Pose Query Encoding to involve the predicted structural pose information into instance feature, which is used to predict a pose regression quality score. Concretely, we convert the general feature $R_{g}$ to produce the raw instance representation $R_{I}$, then concatenate the feature vector at the position of all keypoint queries, as illustrated in Figure \ref{fig:image4}. The above process transforms the raw instance representation $R_{I}$ to generate the new instance representation $R_{I}^{'}$, which is formulated as follows:

\begin{equation}
R_{I}^{'}(c) = Concat(\{{ R}_{I}(c+ \bar{\bf D}^{i}_{c\rightarrow q})\}_{i=1}^{K}),
\end{equation}

\noindent where $K$ is the number of keypoint queries, $\bar{\bf D}^{i}_{c\rightarrow q}$ is the displacement from center $c$ to i-th keypoint query. Due to each keypoint query encode the positional and semantic information for corresponding keypoint. Thus, the Pose Query Encoding is capable of encoding the predicted structural pose information into instance feature representation. Finally, we leverage the new instance representation ${R}_{I}^{'}$ to predict the pixel-wise score map $\bar{\mathcal{I}}$ to better fit the Consistent Instance Representation (CIR).

\section{Experiments}

In this section, we first briefly introduce our experimental setup. Then we carry out the ablation study to investigate the effectiveness of each components of our proposed network. Finally, we conduct the comprehensive comparisons with previous state-of-the-art methods to verify the superiority of our proposed network. 

\subsection{Experimental Setup}

\noindent {\bf Dataset.} We conduct our experiments on widely-used pose estimation benchmark MS COCO \cite{lin2014microsoft}, which includes 200k images with 250k human instance annotated with the positions of 17 body joints. Following previous settings, we leverage coco train2017 with 57k images for training, mini-val set with 5k images for conducting ablation studies, test-dev set with 20k images for comparing with the previous state-of-the-art methods.

\noindent {\bf Evaluation Metric.} The evaluation metrics are average precision and average recall scores based on different Object Keypoint Similarity (OKS) thresholds from 0.5 to 0.95 to evaluate the performance.

\noindent {\bf Augmentation.} In training stage, we carry out data augmentation via random flip with probability of 0.5, random rotation in $[-30,30]$ degrees, random scaling of $[0.75,1.5]$, random shift of $[-40,40]$ pixels as well as color jitter to augment training samples. Each input is cropped to 512 / 640 pixels. The output size is 1/4 of the input resolution. In test process, we use the horizontal flip and multi-scale image pyramids to boost the performance.

\noindent {\bf Implementation Details.} We train our proposed network via Adam optimizer with a mini-batch size of 64. The initial learning rate is set as 5e-4
and dropped to 5e-5 and 5e-6 at the 150th and 170th epochs respectively. Furthermore, the radius of center-neighboring area $\gamma$ is set to 4. The loss weight of ${L}_{\mathcal{I}}$ and
${L}_{{\bf D}_{c\rightarrow k}}$ are both set to 1.0. For inference, we keep the aspect ratio of raw input image and resize the short side of the images to 512 / 640 pixels.

\subsection{Ablative Analysis}\label{subsection4.2}
In this subsection, we first report the contributions of each component in our framework. Then, we delve into the design of them. All ablation studies adopt HRNet-W32 as backbone with single-scale testing on the COCO mini-val set.

\begin{table}
\begin{center}

\resizebox{1.0\columnwidth}{!}{
\begin{tabular}{l|ccc|ccccc}
Expt.& CIR & KQE & PQE  &$AP$ & $AP_{50}$ & $AP_{75}$ & $AP_{M}$ & $AP_{L}$     \\

\hline
1&- & - &-& 63.5 & 85.5 & 69.4 & 56.5 & 73.8  \\
2& $\surd$ & - &-& 64.5 & 85.8 & 71.0 & 58.0 & 74.6  \\
3&$\surd$ & $\surd$ &-& 68.4 & 87.1 & 74.6 & 62.2 & 77.5  \\
4& - & $\surd$ &$\surd$& 68.3 & 86.9 & 75.1 & 62.0 & 77.3  \\

5&$\surd$ & $\surd$ &$\surd$& {\bf 69.8} & {\bf 88.1} & {\bf 76.2} & {\bf 63.8} & {\bf 78.9}  \\

\end{tabular}
}
{\caption{Contributions of each components. CIR denotes the Consistent Instance Representation. KQE is the Keypoint Query Encoding. PQE indicates the Pose Query Encoding.}\label{tab:Contribution}}
\end{center}
\end{table}

\begin{table}
\begin{center}

\resizebox{1.0\columnwidth}{!}{
\begin{tabular}{l|c|cccccc}
Instance Rep. & QEB & $AP$ & $AP_{50}$ & $AP_{75}$ & $AP_{M}$ & $AP_{L}$     \\
\hline

Discrete $\{0,1\}$& - & 67.7 & 86.7 & 73.9 & 61.4 & 77.2  \\
Discrete $\{0,1\}$& $\surd$ & 68.6 & 87.2 & 75.0 & 62.5 & 77.9  \\
Gaussian & - & 68.3 & 86.9 & 75.1 & 62.0 & 77.3   \\
Gaussian& $\surd$ & 69.0  &87.6&75.4  & 63.1 & 78.3 \\
CIR(ours)& - & {\bf 69.8} & {\bf 88.1} & {\bf 76.2} & {\bf63.8} & {\bf78.9}  \\

\end{tabular}
}
{\caption{Comparisons with previous instance confidence representation for denoting the confidence of instance localization. QEB refers to employ an additional quality estimation branch to modulate the instance score.}\label{tab:representation}}
\end{center}
\end{table}

{\bf Contributions of each components.} We analyze the contribution of each component in our proposed method. The results are shown in Table \ref{tab:Contribution}.~Note that we adopt the Gaussian kernel as instance confidence label to denote the person instance in Expt.1, 4. With only Consistent Instance Representation (CIR) applied, we achieve 1.0 AP improvements as reported in Expt.1 and Expt.2. Keypoint Query Encoding (KQE) improves 3.9 AP based on CIR as shown in Expt.2 and Expt.3. Pose Query Encoding (PQE) is able to obtain 1.4 AP improvements in Expt.3 and Expt.5. Furthermore, as shown in Expt.4 and Expt.5, based on the KQE and PQE, the Consistent Instance Representation (CIR) is capable of improving 1.5 AP. The exhaustive analysis for each components will be described in below.


{\bf Analysis of Consistent Instance Representation.} The previous studies \cite{tian2019directpose, nie2019single}
employ discrete $\{1,0\}$ or the 2-dimensional Gaussian kernel to denote the confidence of instance localization. The former assigns the pixels of human central area with label $\{1\}$. This scheme treats each pixel of central area equally while ignores the difference of pose regression quality. The later hypothetically considers that the center position will predict the optimal pose. The farther away from the center, the worse the pose regression quality is. However, both of them are manually settled and may result in the gap between instance score and corresponding pose regression quality.

\begin{figure}
\begin{center}
\includegraphics[height=0.43\columnwidth,width=1.0\columnwidth]{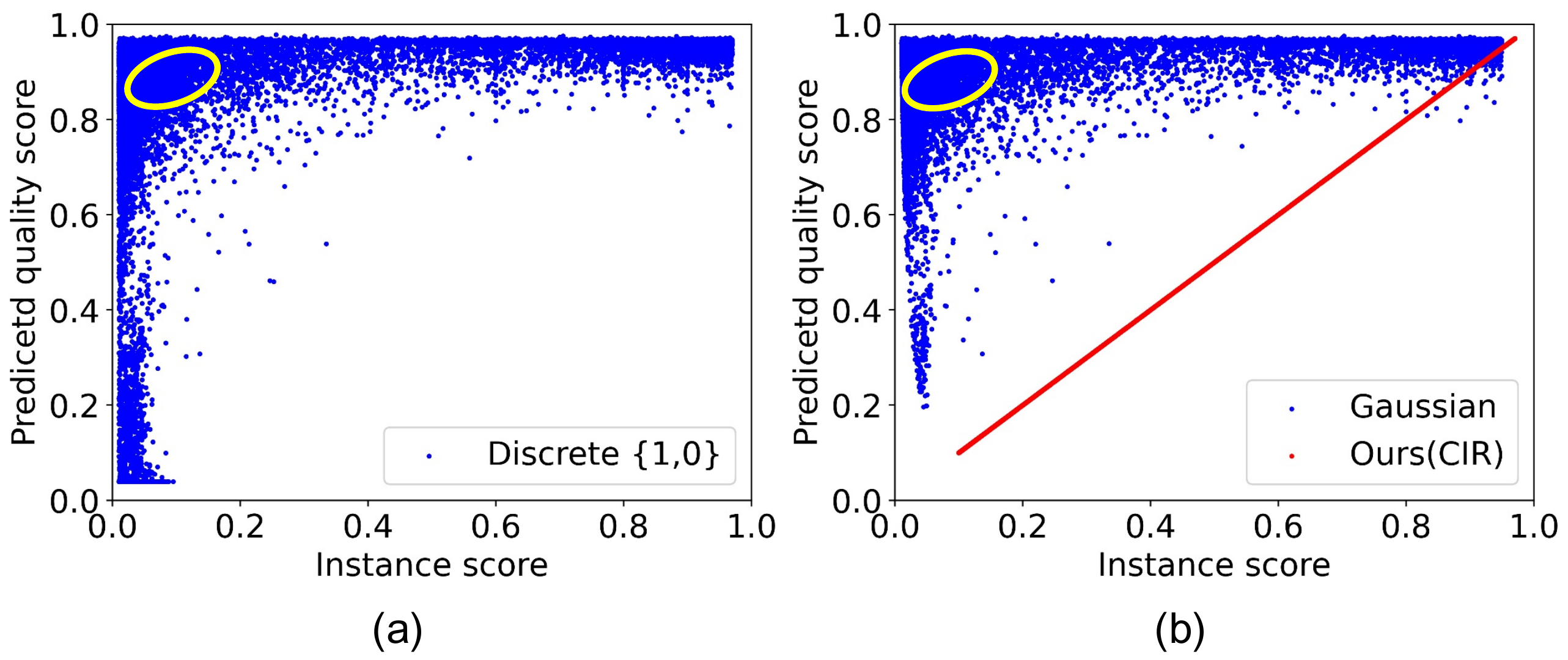}
\end{center}
\caption{The blue points in scatter diagram (a) and (b): employing an additional quality estimation branch to modulate instance score supervised by discrete \{1,0\} or Gaussian representation. The points denote the randomly remained pose candidates after NMS (5 per image) with the predicted instance score and pose quality score. The blue points in yellow circles indicate the pose candidates with low instance score and high pose quality score. Some of these may predicted by background area which lead to false positives, due to the supervision of additional quality estimation is only assigned for instance area during training. The red points in scatter diagram (b): our CIR avoids the unreliable bias caused by multiplying the instance score and uncontrollably high quality score, thus achieves the better performance.}
\label{fig:image6}
\end{figure}

For the previous two instance confidence labels, we further add an additional quality estimation branch, following \cite{jiang2018acquisition,huang2019mask}, whose supervision is only assigned for instance area. The estimated instance score and pose regression quality score are multiplied to select pose candidates during inference. We construct 4 contrasts including discrete \{1,0\} without or with additional quality estimation branch, Gaussian distribution without or with additional quality estimation branch. Our CIR unifies the pose regression quality score of instance area and confidence of background into a single pixel-wise score map thus the additional quality estimation branch is no longer required. As shown in Table \ref{tab:representation}, we observe that the additional quality estimation branch is able to consistently improve the performance for both discrete $\{1,0\}$ and Gaussian representation. However, as shown in Figure \ref{fig:image6}, we observe that employing an additional quality estimation branch to modulate the instance score may lead to false positives (the blue points in yellow circles). Due to the supervision of quality estimation is only assigned for instance area, thus the network may predicts the uncontrollably high score in background to raise the corresponding low instance score. Our proposed Consistent Instance Representation unifies the pose regression quality score of instance area and confidence of background into a single pixel-wise map which avoids the unreliable bias caused by multiplying the instance score and uncontrollably quality score in background. As a result, our CIR achieves the better performance.




\begin{table*}
\begin{center}

\resizebox{1.7\columnwidth}{!}{

\begin{tabular}{l|c|ccccc|c}
\hline
Methods &Input size & $AP$ & $AP_{50}$ & $AP_{75}$ & $AP_{M}$ & $AP_{L}$& $AR$  \\

\hline
\multicolumn{8}{c}{Bottom-up Methods}\\
\hline
CMU-Pose$^{*\dag}$~\cite{cao2017realtime} &-&61.8& 84.9& 67.5& 57.1 &68.2& 66.5   \\
AE$^{*\dag}$~\cite{newell2017associative} &512 & 65.5 & 86.8 & 72.3 & 60.6 & 72.6& 70.2   \\
CenterNet-DLA~\cite{zhou2019objects} &512 &57.9&84.7&63.1&52.5&67.4& -  \\
CenterNet-HG~\cite{zhou2019objects} &512 & 63.0 & 86.8 & 69.6 & 58.9 & 70.4& - \\
PifPaf~\cite{kreiss2019pifpaf} & 801 & 66.7 & - & -& 62.4& 72.9& -   \\
HrHRNet-w48$^{*}$~\cite{cheng2020higherhrnet} &512& 68.4& 88.2 & 75.1 & 64.4 & 74.2& -  \\
SWAHR(HrHRNet-W48)$^{*}$~\cite{luo2021rethinking} &640&70.2&89.9&76.9&65.2&77.0&-\\

\hline
\multicolumn{8}{c}{Single-stage Regression Methods}\\
\hline
SPM $^{* \dag}$~\cite{nie2019single} & - & 66.9& 88.5& 72.9& 62.6& 73.1& -   \\
DirectPose $^{\dag}$~\cite{tian2019directpose} &800*1333 &64.8& 87.8& 71.1& 60.4& 71.5& - \\
MDN$_{3}$ ~\cite{varamesh2020mixture} & - & 62.9& 85.1& 69.4 & 58.8& 71.4& -  \\
PointSetNet $^{* \dag}$~\cite{wei2020point} &640 &68.7& 89.9& 76.3& 64.8& 75.3& 74.8  \\
DEkR-W48$^{*}$\cite{geng2021bottom} & 640 & 70.0 & 89.4 & 77.3 & 65.7&76.9&75.4 \\ 
DEkR-W48 $^{* \dag}$\cite{geng2021bottom} & 640 & 71.0 & 89.2 & 78.0 & 67.1&76.9&{\bf76.7} \\ 
\hline

Ours~(HRNet-W32) &512& 69.0 & 89.3 & 76.0 & 62.8 & 77.0&73.6     \\
Ours~(HRNet-W32)$^{\dag}$ &512& 70.5 & 89.6 & 77.5 & 64.8 & 78.0&75.1   \\
Ours~(HRNet-W48) &640& 71.0 & 90.2 & 78.2 & 66.2 & 77.8 & 76.0 \\
\bf{Ours~(HRNet-W48)}$^{\dag}$ &640& {\bf71.7} & {\bf90.4} & {\bf78.7} & {\bf67.3} & {\bf78.5} &76.5   \\
\hline
\end{tabular}
}
{\caption{Comparisons with the state-of-the-art methods on COCO test-dev set. ${*}$ indicates using additional post-process(e.g., single-person model refinement used in CMU-Pose, AE, SPM and pose scoring net in DEKR). $\dag$ refers to multi-scale testing.}\label{tab:test}}
\end{center}
\end{table*}


\begin{table}
\begin{center}

\resizebox{0.8\columnwidth}{!}{
\begin{tabular}{l|cccccc}
N & 0 & 3 & 6 & 9 & 12 & 15     \\
\hline
AP &  68.5 & 68.9 & 69.2 & {\bf69.8} & 69.6& 69.1   \\
\end{tabular}
}
{\caption{Ablation study for varying the number of semantic points N by fixing the other proposed components.}\label{tab:N}}
\end{center}
\end{table}

\begin{table}
\begin{center}

\resizebox{0.8\columnwidth}{!}{
\begin{tabular}{l|cccccc}
Method & $AP$ & $AP_{50}$ & $AP_{75}$ & $AP_{M}$ & $AP_{L}$     \\
\hline

Auto & 68.8 & 87.6 & 75.1 & 62.3 & 77.9  \\
SP &  68.9 & 87.8 & 75.0 & 62.5 & 78.3   \\
KPS &  69.0   &87.5&75.1  & 62.6 & 78.5  \\
KQ (ours)&  {\bf69.8} & {\bf88.1} & {\bf76.2} & {\bf63.8} & {\bf78.9}  \\

\end{tabular}
}
{\caption{Ablation study for the construction of pose query. Auto: aggregating the features via the positions of 17 automatically located points; SP: aggregating the features via the positions of 17*N semantic points; KPS: aggregating the features via the position of 17 keypoints; KQ: aggregating the features via the position of 17 keypoint queries. }\label{tab:pose-query}}
\end{center}
\end{table}


{\bf Analysis of Keypoint Query Encoding.} The semantic information for each keypoint query is supplemented via the features of N semantic points. We explore the influence for the number of semantic points in Keypoint Query Encoding. As shown in Table \ref{tab:N}, our method achieves the better performance with the number of semantic point increasing. It proves that enriching the semantic feature for keypoint query is able to more precisely regress the keypoint. We achieve the best performance with 69.8 AP when N is set to 9.

{\bf Analysis of Pose Query Encoding.} Pose Query Encoding aims to encode the predicted structural pose information into instance feature to predict pose regression quality score. Based on the other proposed components, we investigate how to generate the pose query that better involve the predicted pose information into instance feature. As reported in Table \ref{tab:pose-query}, we construct 4 contrasts for comparisons, the results prove that aggregating the features via the positions of 17 keypoint queries to construct the pose query achieves the better performance compared with the others.


\subsection{Comparison with the State-of-the-art Methods}\label{subsection4.3}

We compare our method with the current multi-person pose estimation methods on COCO mini-val and test-dev set.


{\textbf{Mini-val Results.}} Table~\ref{tab:val} reports the performance of single-scale testing on COCO mini-val set. With HRNet-W32 as backbone, our method achieves 69.8 AP when input resolution is set as 512 pixels and outperforms the previous bottom-up methods   \cite{papandreou2018personlab,zhou2019objects} with a large margin. In particular, compared with the state-of-the-art HigherHRNet \cite{cheng2020higherhrnet} and DEKR\cite{geng2021bottom}, our network achieves 2.7 AP and 1.8 AP improvements without either multi-scale heatmap aggregation or additional pose scoring net during inference. We further obtain 72.4 AP with input resolution of 640 pixels via HRNet-W48, which is a new state-of-the-art performance compared with all existing single-stage as well as bottom-up methods.

{\textbf{Test-dev Results.}} We compare our approach with the prior state-of-the-art bottom-up and single-stage regression-based methods on test-dev2017 set. The results are reported in Table \ref{tab:test}. Adopting HRNet-W48 as backbone with single-scale testing, our method achieves 71.0 AP which outperforms the bottom-up AE \cite{newell2017associative}, CenterNet-HG \cite{zhou2019objects} as well as PifPaf \cite{kreiss2019pifpaf} with a large margin, and surpasses the state-of-the-art HigherHRNet-W48 + AE \cite{cheng2020higherhrnet} and SWAHR-W48 \cite{luo2021rethinking} by 2.6 AP and 0.8 AP respectively. Compared with single-stage regression-based methods, our approach achieves 4.8 AP gains over SPM \cite{nie2019single}, 6.9 AP gains over Directpose \cite{tian2019directpose} and 3.0 AP improvements over PointsetNet \cite{wei2020point}. Moreover, we improve 1.0 AP and 0.7 AP in comparison to state-of-the-art regression-based model DEKR-W48 \cite{geng2021bottom} for single-scale and multi-scale testing.


\begin{figure*}
\begin{center}
\includegraphics[height=0.735\columnwidth,width=2.0\columnwidth]{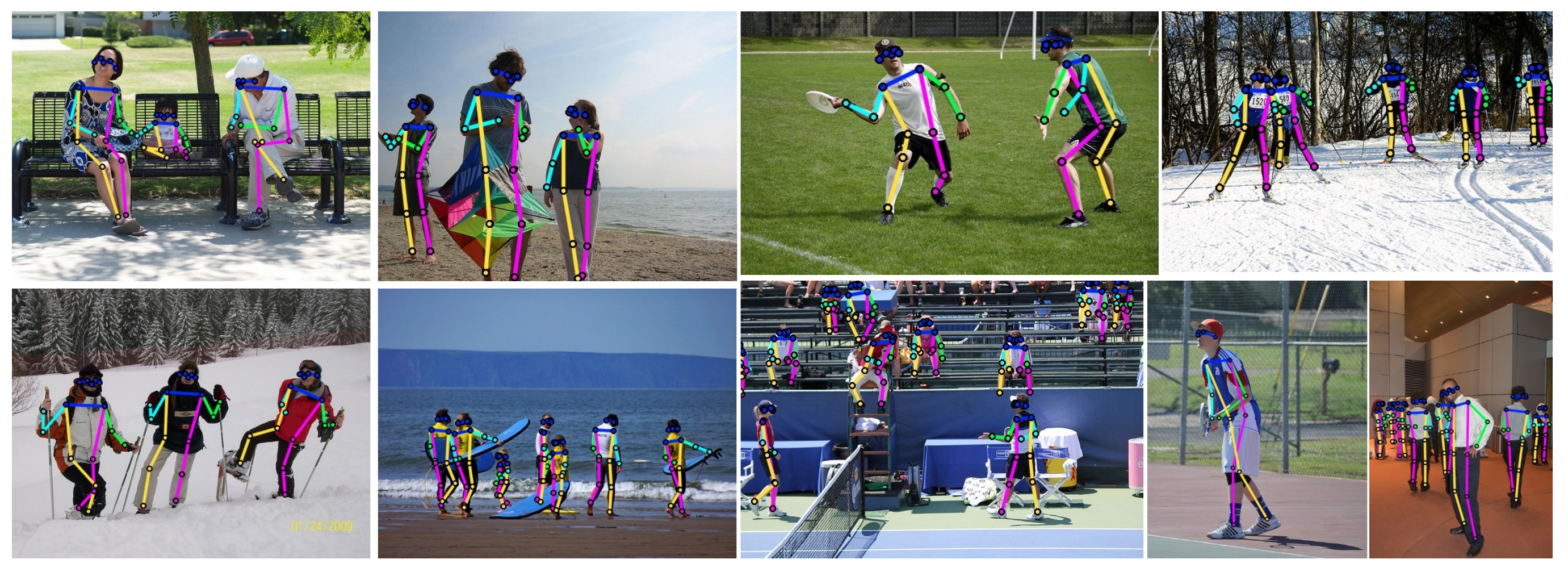}
\end{center}
\caption{Examples of predicted poses for diversity scenarios on COCO dataset.}
\label{fig:image7}
\end{figure*}


\section{Conclusion}

In this paper, we propose to learn the pose regression quality-aware representation. Concretely, we firstly present a Consistent Instance Representation (CIR) that unifies the pose regression quality score of instance area and presence confidence of background into a single pixel-wise score map to fill the inconsistency between the instance score and pose regression quality. Furthermore, we present a Query Encoding Module (QEM) that encodes the keypoint information for precisely regression and involves the predicted structural pose information into instance feature representation to predict the reasonable pose regression quality score. Based on the proposed CIR and QEM, our network is able to significantly alleviate the above gaps existing in current single-stage pose regression practices. Comprehensive experiments demonstrate the state-of-the-art performance of our proposed method.

\section{Acknowledgements}

This work is supported by the National Natural Science Foundation of China (62071056).

\end{document}